# An Artificial Immune System as a Recommender for Web Sites

**Tom Morrison**
University of the West of England
Frenchay Campus Bristol
BS16 1QY
tom.morrison@uwe.ac.uk

**Uwe Aickelin**
School of Computer Science
University of Nottingham
NG8 1BB UK
uxa@cs.nott.ac.uk

**Abstract**

Artificial Immune Systems have been used successfully to build recommender systems for film databases. In this research, an attempt is made to extend this idea to web site recommendation. A collection of more than 1000 individuals' web profiles (alternatively called preferences / favourites / bookmarks file) will be used. URLs will be classified using the DMOZ (Directory Mozilla) database of the Open Directory Project as our ontology. This will then be used as the data for the Artificial Immune Systems rather than the actual addresses. The first attempt will involve using a simple classification code number coupled with the number of pages within that classification code. However, this implementation does not make use of the hierarchical tree-like structure of DMOZ. Consideration will then be given to the construction of a similarity measure for web profiles that makes use of this hierarchical information to build a better-informed Artificial Immune System.

## 1 INTRODUCTION

This research is concerned with using Artificial Immune Systems as a recommender of web sites for new database members. Thus, a new member of the database system would be able to export their bookmark / favourites file and receive a small number of recommendations of web site addresses (URLs or Uniform Resource Locators). Unlike a search engine that will only return specific items a user searches for, our recommender system should be capable of providing the user with surprising items of interest.

Artificial Immune Systems are adaptive search algorithms based on the biological immune system with the central task of pattern matching between antigens and antibodies. Thus in our opinion, they are particularly well suited to data-mining tasks that involve sifting through large databases and finding matches to other items. This has been confirmed in recent research by Cayzer and Aickelin [5] who used Artificial Immune Systems to recommend films to new members of a database based on their rating of at least five films.

As in the research by Cayzer and Aickelin, the type of Artificial Immune System developed here will be based on Jerne's idiotypic network ideas [13]. Hence, we will build an Artificial Immune System that will find a group of users in the database who are similar to the target user in their web site preferences. At the same time, the idiotypic effects will ensure that this group is as diverse as possible. Thus, we will have created an ideal base for predicting and recommending web sites. To do this successfully two steps are necessary: building a database that models individuals' web profiles using a suitable ontology, and constructing a suitable measure of how similar two web profiles are.

The remainder of this paper is organised as follows: In the next section, a very brief overview of the immune system is given with particular emphasis on those features that we intend to exploit here. Section 3 will summarise the research into film prediction and explain differences and similarities to this piece of research. The following section describes the data and ontology used and gives further details about the task of web site recommendation. Section 5 presents a description of the intended Artificial Immune System with an emphasis on the discussion of a suitable similarity measure. The paper is concluded with a summary.

## 2 THE IMMUNE SYSTEM

The human body is protected against foreign invaders by a multi-layered immune system. The immune system is composed of physical barriers such as the skin and respiratory system; physiological barriers such as destructive enzymes and stomach acids; and the immune system, which has two complementary parts, the innate and adaptive immune systems. The innate immune system is an unchanging mechanism that detects and destroys certain invading organisms, whilst the adaptive immune system responds to previously unmet foreign cells and builds a response to them that can remain in the body over time.

The immune system is composed of a number of different agents performing different functions at a number of different locations in the body. The precise interaction of these agents is still a topic for debate [10]. In order to present the important aspects of the system from a mathematical viewpoint it is necessary to simplify and present a selective description.

The immune system's job is to detect antigens, which are foreign molecules from a bacterium or similar invader. The innate immune system helps in the detection process but the main response is through the adaptive immune system. Two of the most important cells in this process are white blood cells, called T cells, and B cells. Both of these originate in the bone marrow but T cells pass on to the thymus to develop before, as with B cells, they circulate the body in the blood and lymphatic vessels.

B cells are responsible for the production and secretion of antibodies, which are specific proteins that bind to the antigen. Each B cell can only produce one particular antibody. The antigen is found on the surface of the invading organism and the binding of an antibody to the antigen is a signal to destroy the invading cell. A diagram from de Castro and Von Zuben [4] of this process is shown in Figure 1.

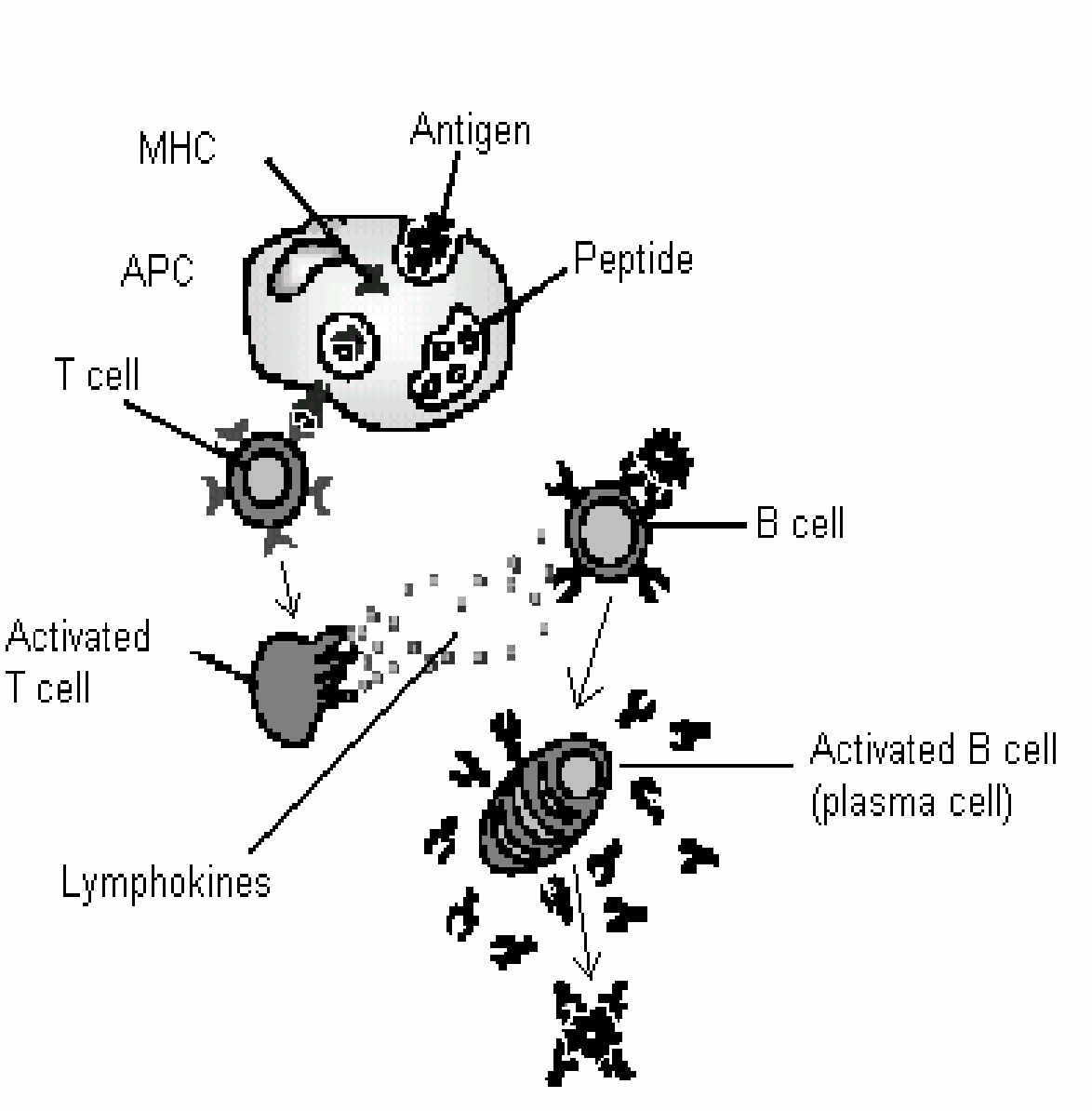

Figure 1: Some of the processes involved in the adaptive immune system.

Whilst there is more than one mechanism at work (see [8], [10] or [15] for more details), the essential process for the sake of this research is the matching of antigen and antibody leading to increased concentrations of more closely matched antibodies. In particular, two processes, known as the 'clonal selection theory' by Burnet [3] and the 'idiotypic network theory' by Jerne [13] and [14], are important to us.

The former can be explained as follows: When an antibody strongly matches an antigen the corresponding B cell is stimulated to produce clones of itself that then produce more antibodies. This selection of B cells for cloning on the basis of the antibody match is called the 'clonal selection principle' and will result in increasing concentrations of that antibody in the body.

However, when the B cells clone themselves they do not do so exactly, but mutate slightly. Similarly, B cells may be stimulated when the antibody-antigen match is not perfect. By allowing mutation, the match could become better. However, a number of poorer matches will also be created, and furthermore, some of the newly produced antibodies could even be harmful to our own cells. Such cells will die out under what is known as the 'negative selection principle' [10].

The mutation, mentioned above, is quite rapid, often as much as de Castro and Von Zuben state in [4] "one mutation per cell division". This allows a very quick response to the antigens. This rapid mutation, known as 'somatic hypermutation' [10], may be linked to the 'fitness' of the antibody. Hence, those B cells producing antibodies that are a good match would be subject to less mutation and vice versa for those that are not such a good match.

The idiotypic network theory, introduced by Jerne in [13] and [14], maintains that interactions in the immune system do not just occur between antibodies and antigens, but that antibodies may interact with each other. Hence, an antibody may be matched by other antibodies, which in turn may be matched by yet other antibodies. This activation can continue to spread through the population. However, this interaction can have positive or negative effects on a particular antibody-producing cell. The idiotypic network has been formalised by a number of theoretical immunologists in [15]. This theory could help explain how the memory of past infections is maintained. Furthermore, it could result in the suppression of similar antibodies thus encouraging diversity in the antibody pool.

This last possibility was used in the research by Cayzer and Aickelin [5] in order to preserve diversity. The Artificial Immune System in their research produced a pool of users who were similar to the new entrant to the database, but dissimilar to each other. Whilst this method produced similar performance in predicting film ratings to a k-nearest neighbour approach, the diversity in the pool of recommenders was found to yield statistically significantly improved recommendations. Given the sparseness of the web site search space it may be that suppression of antibodies on similarity grounds might be unnecessary. This will be investigated.

There are a number of successful Artificial Immune System implementations. However, even in the most complex artificial systems only a fraction of the functionality of the biological immune system is exploited. Typically, the antibody-antigen interaction coupled with somatic hypermutation, form the basis for many Artificial Immune System applications. Examples are Timmis et al [18], who used an Artificial Immune System for clustering multivariate data, and Hajela and Yoo [11], who combined a genetic algorithm and an Artificial Immune System to optimise the design of a 10 bar truss. The research by Timmis et al also applied the idiotypic network theory and were successful in both classifying data and "generalising to cover a larger region of the input space". However, the article does not comment on the effect of modelling a suppression factor between antibodies. Some of the most promising research to date has been conducted in the area of computer security, for instance by Hofmeyr and Forrest in computer network security [12] and by Kim and Bentley for fraud detection [15] and [16].

## 3 ARTIFICIAL IMMUNE SYSTEMS AS RECOMMENDERS

Whilst most of the applications described above involve somatic hypermutation, Cayzer and Aickelin [5] had only identical cloning, not mutation, in their algorithm. This was because the potential antibodies were actual users of the film database (EachMovie database provided by the Compaq Research Centre [6]). There the task was to find users that were similar to new entrants to the database. Somatic hypermutation was not used, since it is not immediately obvious how to mutate users sensibly such that these artificial entities still represent plausible profiles.

For the same reasons, cloning in our intended Artificial Immune System will make exact copies, too. Future work might include making inexact copies to create novel profiles once appropriate rules for doing so have been established. This could be particularly beneficial when data gathering is expensive or data is otherwise sparse, perhaps due to its sensitive nature, leading to few users being willing to share their information with others.

The main loop of the recommender algorithm is shown in Figure 2 below and is the core of our Artificial Immune System. The aim of this algorithm is to increase the concentrations of those antibodies (database users) that are similar to the antigen (target user). This process is subject to the suppression of similar antibodies following Jerne's idiotypic ideas mentioned above. Thus, over time the Artificial Immune System contains high concentrations of a diverse set of users who have similar film preferences to the target user.

```
Initialise AIS
Encode user for whom to make predictions as antigen Ag
WHILE (AIS not stabilised) & (More data available) DO
  Add next user as an antibody Ab
  Calculate matching score between Ab and Ag
  Calculate matching scores between Ab and antibodies
  WHILE (AIS at full size) & (AIS not stable) DO
    Iterate AIS
  OD
OD
```

Figure 2: Main loop of the Artificial Immune System's (AIS) algorithm for recommendation.

The diagrams in Figure 3 show the idiotypic effect. In the top diagram, antibodies $Ab_1$ and $Ab_3$ are very similar and they would have their concentrations reduced in the 'Iterate AIS' stage of the algorithm above. However, in the lower diagram, the four antibodies are well separated from each other as well as being close to the antigen and so would have their concentrations increased.

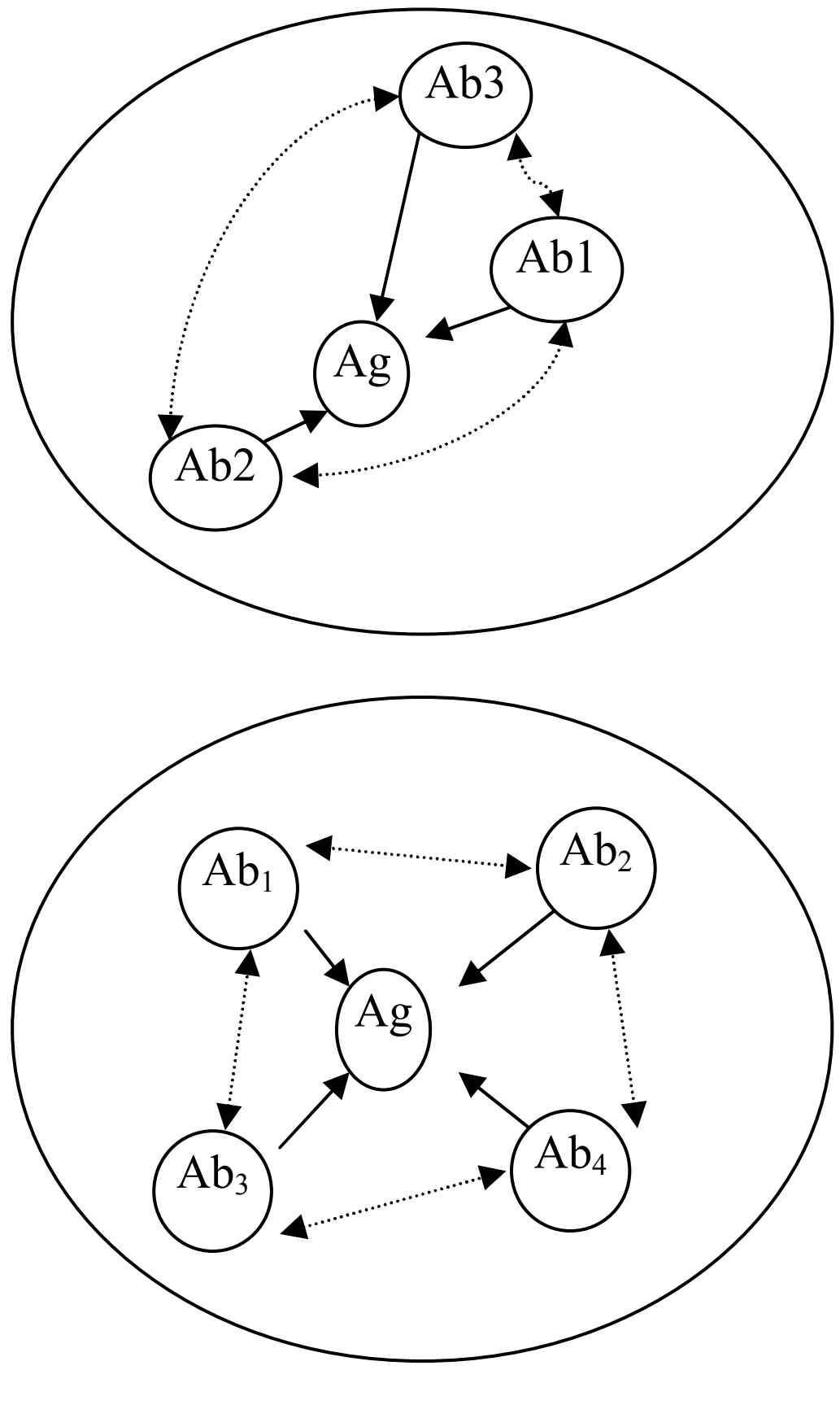

Figure 3: Illustration of the idiotypic effect.

At each iteration of the film recommendation Artificial Immune System the concentration of the antibodies changes according to the formula given below. This will increase the concentration of antibodies that are similar to the antigen and can allow either the stimulation, suppression, or both, of antibody-antibody interactions to have an effect on the antibody concentration. More detailed discussion of these effects on recommendation problems are contained within Cayzer and Aickelin's paper [5].

The following is a formal equation for the idiotypic effect adapted from Equation 3 in Farmer [8]:

$$\frac{dx_i}{dt} = c\left[\begin{pmatrix} antibodies \\ recognised \end{pmatrix} - \begin{pmatrix} I\ am \\ recognised \end{pmatrix} + \begin{pmatrix} antigens \\ recognised \end{pmatrix}\right] - \begin{pmatrix} death \\ rate \end{pmatrix}$$

$$= c\left[k_0 \sum_{j=1}^{N} m_{ji} x_i x_j - k_1 \sum_{j=1}^{N} m_{ij} x_i x_j + k_2 \sum_{j=1}^{N} m_{ji} x_i y\right] - k_3 x_i$$

Where:

$N$ is the number of antibodies
$x_i$ is the concentration of antibody $i$
$m_i$ is the antibody $i$ and the antigen correlation
$m_{ij}$ is the correlation between antibodies $i$ and $j$
$y$ is the concentration of the antigen
$k_1$ is suppression, $k_2$ stimulation and $k_3$ death rate
$k_0$ is set to zero in our system, i.e. we do not reward antibody - antibody recognition.

The algorithm is terminated, when the Artificial Immune System is said to have stabilised, i.e. if it has not changed in consistency for more than ten iterations. The concentrations and correlations of the users in the final neighbourhood, i.e. final immune system iteration, are then used to calculate a weighted sum of the ratings of web sites. This would be either a specific unseen web site by the target user in order to predict its ratings, or general top 10 recommendations of new web sites that the target user might enjoy.

# 4 THE CHALLENGE OF WEB SITE RECOMMENDATION

There are a number of algorithms that recommend items to users. One of the best-known examples is Amazon.com's [1] book recommender based on similar items bought. Generally, these recommenders use what is termed "collaborative filtering" or "social filtering" by Billsus and Pazzani [2]. With the exponential growth of available information on the internet, the need for automated techniques to winnow down the possibilities has also grown but "only a few different algorithms have been proposed in the literature thus far" [2].

Many of the current collaborative filtering techniques use the Pearson correlation coefficient to compare the item ratings of different users. This suffers from several limitations. For example, due to the extremely large amount of information to be rated, two users may only have a very small number of items in common causing the correlation measure to be unduly influenced by those items. Further, there is potentially no difference between the correlation between two users with three items in common and the measure for two users with 30 items in common, in terms of their "influence on the final prediction" [2].

The sparseness of the information space also implies that two users might have no items in common. Can we therefore conclude that they have completely dissimilar tastes, or does the fact that they have not rated particular items imply a similar view of the importance of those items? For these reasons, alternative approaches to both current collaborative filtering algorithms and to the use of the Pearson correlation coefficient should be investigated. More information about traditional and enhanced collaborative filtering is provided by Gokhale [9]. The Artificial Immune System presented here is another example.

In our problem of web site recommendation, the original data consists of sets of web site addresses or URLs taken from bookmark collections such as http://www.cs.ucl.ac.uk/staff/Kim/ComputerImmune. It is extremely unlikely that many people will have many exact addresses in common within their web profiles. Because of this, it is necessary to transform or translate the addresses into a different form. To do this a number of steps are necessary and a widely used web site classification tree ontology will be used called DMOZ [7].

Let us look at the issues involved in the classification of URLs systematically. Typically, an individual web profile in raw form might consist of a list of bookmarks as shown in Figure 4 (in this case taken from the Opera browser – only a small section is shown).

```
#URL
    NAME=ODP - Open Directory Project
    URL=http://dmoz.org/
    CREATED=1017158736
    VISITED=1023875733
#URL
    NAME=Open Directory RDF Dump
    URL=http://dmoz.org/rdf.html
    CREATED=1017159133
    VISITED=1023875759
```

Figure 4: Part of a raw web profile taken from the Opera browser.

This data has to be pre-processed in order to remove unwanted information and superfluous characters. This also includes removing any categories the user might have assigned to some of the bookmarks. Unfortunately, such categorisation of information cannot be kept, as it is arbitrary and individual to the person that owns the bookmarks. For instance, www.bbc.co.uk could be classified under 'media' by one person and under 'news' by another. In addition, misclassifications and duplications might be present in the raw data. Hence, this filtering typically yields a file such as the one partially shown in Figure 5.

```
www.bbc.co.uk/weather/
www.bbc.co.uk/
www.bbc.co.uk/sport/english/football/default.stm
www.guardian.co.uk/
football.guardian.co.uk/
```

Figure 5: Part processed data with superfluous information deleted.

As can be seen from the third line in Figure 5, some of the URLs will have long addresses. Another web profile might contain a very similar address such as *www.bbc.co.uk/sport/english/football/en/default.stm*. If we were to use the raw addresses within the Artificial Immune System, these two would be considered different. However, it is clear that the two users have bookmarked different pages within the same part of the same site, i.e. ‘BBC online - football’, and thus have very similar interests.

Therefore, it is still necessary to process the data before it can be used. This presents considerable problems. A program will need to be devised which will truncate the URLs in such a way so that the two addresses discussed above would be considered the same. However, looking again at Figure 4, a simple truncation of the addresses would lead to the first three items occupying the same category. At the same time, it might not lead to the last two being picked together despite the fact that both the addresses refer to pages from the same site. Furthermore, it might not put items 3 and 5 together despite the fact that they are both concerned with football.

To overcome these difficulties, two strategies are used within the DMOZ ontology: Normalisation and reverse partial look-up. First, all URLs undergo a kind of normalisation when pre-formatting the data, as well as when doing look-ups. The protocol and host part are mapped to lowercase characters and host only URLs are always terminated with a “/”. During the actual look-up, the category information is gained from DMOZ by employing a reverse truncation search. That is, at first, we try to match the full URL, and then we try to match up to the last “/”, then to the last but one “/” etc.

For instance, we would first try to match item three from above by looking for the full URL in DMOZ. If we cannot find that, we would look for www.bbc.co.uk/sport/english/football/; if this fails, we would search for www.bbc.co.uk/sport/english/ etc. Alternatively, we could try to find the closest match in DMOZ defined by the number of consecutive characters that are identical counted from the beginning of the URL.

These normalisation and intelligent matching together should overcome the first problem mentioned above. To overcome problems of misclassification and to have a common standard we decided to use the DMOZ open directory ontology as a classification system [7]. Figure 6 shows part of the structure of this directory.

```
<Topic r:ID="Top/Arts">
<tag catid="2"/>
<d:Title>Arts</d:Title>
<narrow r:resource="Top/Arts/Books"/>
<narrow r:resource="Top/Arts/Music"/>
<narrow r:resource="Top/Arts/Television"/>
[…]
<Topic r:ID="Top/Kids_and_Teens/Pre-School">
<catid>468769</catid>
<link r:resource="http://www.coolplays.com/"/>
<link r:resource="http://kayleigh.tierranet.com/"/>
<link r:resource="http://www.megafile.com.br/"/>
<ExternalPage about="http://www.coolplays.com/">
<d:Title>Coolplay's Cool for Kids</d:Title>
<d:Description>Includes animated nursery rhymes, crafts, alphabet and spelling games, and colouring book.
```

Figure 6: Part of the DMOZ open directory structure.

The first half of Figure 6 shows part of the ‘Arts’ category, which is located immediately below the root of the tree (called Top). Each category has a unique identifier number (2 in this case). This category has a number of sub categories that in turn have several sub categories of their own. In total, there are some 5 million URLs in 428,590 categories spread over 16 levels in the directory. Categories can also be referred to using an address showing the parent categories in a way that preserves the tree structure information. For example, a category address might read ‘1.3.9’ meaning that it is the ninth sub category of category 3, which is the third sub category of category 1.

The second half of Figure 6 shows how URLs are represented in DMOZ and gives an example of a more detailed description of one URL as provided by an anonymous referee. The complete DMOZ database is roughly one GB in size and updated regularly. All specifications in this paper refer to DMOZ as of 1 June 2002. Overall, the version of DMOZ that we use has the following tree structure with deepest branch being 16 levels below the top:

```
 1
 18 /
621 //
6675 ///
30754 ////
61042 /////
68901 //////
101567 ///////
82802 ////////
51454 /////////
20592 //////////
3467 ///////////
616 ////////////
 69 /////////////
 8 //////////////
 2 ///////////////
 1 ////////////////
```

Figure 7: Full DMOZ structural tree.

The final stage of processing the data is to turn each of the URLs, shown in Figure 7, into a file containing either the category identification numbers or the category addresses, coupled with the number of items in each category. The choice about which version to use will be discussed in the next section.

There are a number of possible pitfalls with this process. For example, many profiles will contain a set of URLs, which are created by the browser program that they use. Few users are likely to delete all of these links, reasoning that they may be useful at some stage. This may create a situation of artificial similarity between users, which would prevent the Artificial Immune System from functioning effectively.

Secondly, the process of placing URLs into categories is likely to involve some truncation if at first there is no clear category involved. This could lead to several subtly different addresses being classified into the same category due to the truncation look-up. Depending on whether the truncated sites are from genuinely different URLs or not this could be good or bad. In the first case, the category may appear to be more popular than it should be whereas in the second case the number in the category is a clear indication of interest in that category. Until the data is fully assembled and individual examples are checked, it will not be possible to judge how critical some of these problems will be.

# 5 BUILDING THE ARTIFICIAL IMMUNE SYSTEM RECOMMENDER

In the film recommender research described in Cayzer and Aickelin [5], each user was coded as a user identification number followed by pairs of film identification numbers with the corresponding rating of the film. The target user became the antigen, whilst the current database members were potential antibodies. In each iteration, antibodies were added to the Artificial Immune System. Those judged to be more similar to the antigen in their film ratings had their concentration increased.

A unique feature of that particular approach was the application of the idiotypic network theory by Jerne [13]. This was implemented such that antibodies that were very similar to each other had their concentration reduced. This has the effect of creating a set of users who are similar to the new user but quite different to each other and thus enhancing the recommendation accuracy of the system. We intend to use the same mechanism for our web site recommender to build an Artificial Immune System as described in section 3.

In order to do this, we also have to decide on the encoding of a user's web profile for which there are two possibilities. In both cases, a user is encoded as a list of category IDs and the number of bookmarks within each category. The difference is in the category IDs; they can be either an integer or a reference to the tree structure. To illustrate the difference, Figure 8 shows the same user's bookmarks for both encodings. The figures in bold indicate how many bookmarks fall into a particular category:

Encoding with the Tree structure:
1.13.12.1.5:**5**;
1.13.12.1.6:**3**;
1.16.3.2.11.5:**1**;
1.18.1.2:**1**;

Encoding with integer category IDs:
22343:**5**;
495771:**3**;
334921:**1**;
3409:**1**;

Figure 8: Integer versus Tree Encoding.

If the second encoding is used together with the number of sites within each category as a rating of the popularity of that category then the problem becomes similar to the film recommendation problem.

However, here we have a considerably sparser search space. In the film database, there were approximately 20,000 entries whereas in the DMOZ directory there are over 400,000 categories. This sparseness may prevent the system from working since many users might have nothing in common, or, at best some categories that are common to the vast majority of the data. Furthermore, many users will have only one entry in a number of categories, leading to increased similarity since the 'rating' of that category will be the same. These problems may prevent an Artificial Immune System based on this encoding being successful in identifying a group of similar users.

There is another problem with using integer category IDs. Because DMOZ is an evolving classification system, new categories are added and removed regularly. This can have the effect that two very similar categories end up with very different integer IDs as these are handed out consecutively. For instance, Star Wars part four might have ID 20,004 when it was classified years ago, but Star Wars part two might end up with ID 420,012 because it has only recently entered the DMOZ system. A similar effect can be seen in Figure 8 for the first two bookmarks. Figure 8 also shows how the tree structure IDs might prevent some of these problems as similar categories still end up near each other in the tree.

The alternative to the integer encoding is to use an encoding that includes the tree structure in the form of a category address. What is required then is a similarity measure that carefully recognises categories that are 'close' within the structure of the tree. For example, it would need to judge the parent / child or the sibling relationship as being more similar than a first cousin or grandparent type relationship. However, constructing such a measure is far from simple. Consider the two trees in Figure 9.

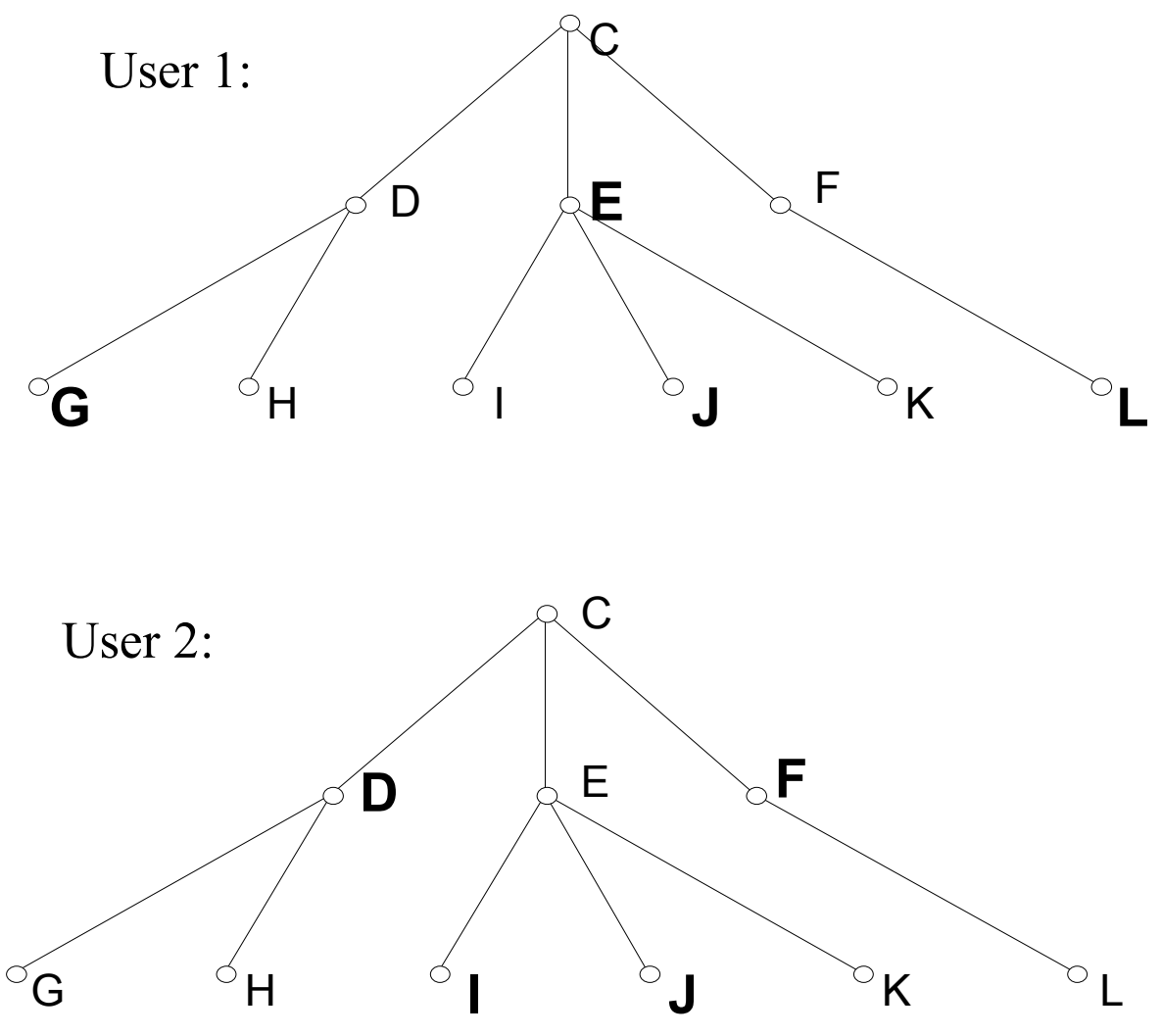

Figure 9: Simple tree structure showing two web profiles.

User 1 has entries at categories G, E, J and L, whilst user 2 has entries at D, I, J and F. Clearly, matches should be scored more highly the lower down the tree they are because this indicates a more precise match. Additionally, ‘close’ relationships within the tree structure should count more towards the match than ones separated by several ‘generations’ (to continue the family tree metaphor).

Whilst it is easy to see that these users should have their similarity measure increased, since both have an entry in category J, a question remains what to do with J afterwards. Should this match be discarded once it has been counted by the measure or should the entries at I and J for user 2 be counted as two entries at the parent branch (E) for comparison with user 1? The danger with discarding matches once counted is that two users might have ‘perfect’ matches for all of the 10 categories that the first user has in their profile, whilst the second user has another 100 entries.

However, if one does not discard categories that have already been matched with another category then it is possible that one quite high level category might be ‘matched’ with all the different entries at sub-categories for another user. This might not matter since the ‘strength’ of the match would have been reduced by the generational distance and the weakness of the high-level category’s contribution.

# 6 SIMILARITY MEASURES

Let us now construct a suitable similarity measure for the Artificial Immune System that will produce a value on a 0–1 scale with answers closer to 1 indicating a closer match. Following the discussion in the previous section, the measure will be built according to the following five principles.

1. Matching at categories lower down the tree structure should contribute more to the measure than matching higher up.
2. Matches at the top level of the tree (i.e. the ‘Top’ category in the DMOZ database should have a contribution of zero.
3. Matching contribution should be reduced for ‘imperfect matches’ i.e. those not in exactly the same category. The reduction in contribution should be proportional to the generational distance (i.e. a grandparent child relationship has a generational distance of two.)
4. The matching metric should be scaled (averaged) so that it ranges from 0 to 1.
5. The matching metric should take into account all possible matches between the entries in each web profile, i.e. if there are 10 entries in 1 and 20 in the other then all 10 × 20 = 200 potential matches should contribute to the measure.

Suppose that we wish to calculate the matching coefficient for the category addresses 1.3.1.1 and 1.3 in the sample tree diagram in Figure 10 below. We need to define an ‘edge distance’ as the number of ‘steps’ apart any two addresses are. For example, 1.1 and 1.1.2.2.1 have an edge distance of three, as do 1.2.2.2 and 1.2.1. This equates the relationship between grandparent and grandchild as the same strength as that between siblings.

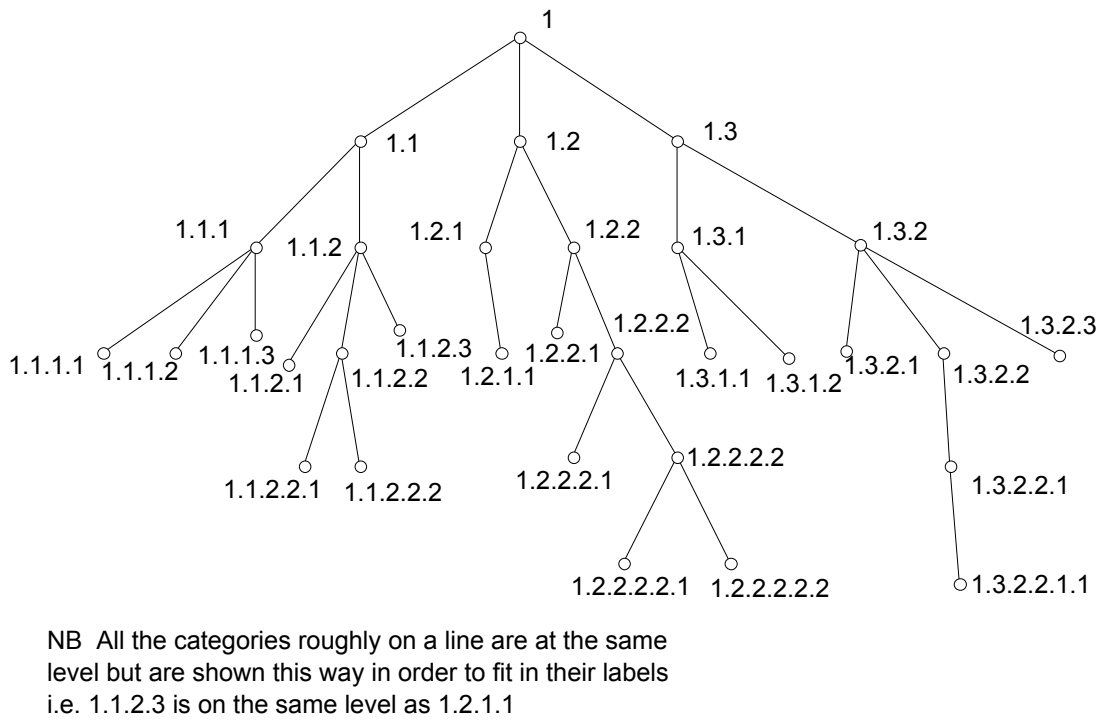

Figure 10: Sample Tree diagram.

By staged truncation of the longer category address (CA) until they are the same we obtain a match at CA 1.3 with two numbers (edge distances) discarded (but counted). This match would have a strength determined by the category level (level 2) of the matching CA, and by the edge distance (ED).

How should the edge distance affect the value of the overall match? One possibility would be to use 1 / ED as this would be a smaller value as the ED increases. However, this would not work when the CA match perfectly as we would be dividing by zero. Therefore using 1 / (ED + 1) is better.

How should the depth of the matching level affect the value of the overall match? It seems useful to make the level number the same as the number of integers in the CA. In the example above, there are six levels. However, the tree is not of uniform depth. In principle, matches at lower levels should score higher since they show a more precise agreement in the topic matter. However, does this mean that a perfect match at the bottom of one set of branches (e.g. 1.1.2.2.2) should score less highly than a perfect match at the bottom of another lower set, say 1.3.2.2.1.1? The DMOZ database is a human classification of human knowledge. To some extent, the classifications are arbitrary because they are the result of pragmatic as well as epistemological considerations. Therefore, it seems incorrect to allow only a perfect match score when it occurs at the lowest level.

In the example above it might be advisable to allow perfect matches to contribute fully at levels 4,5 and 6. Remembering that a match at the top level should count as zero then a formula to give the level effect factor would be (L - 1) / (4 - 1) i.e. level 4 would have a value of 1, level 3 a value of (2/3), level 2 (1/3), whilst the top level would have a value of zero. However, this would not work for values of L greater than 4. To solve this we could use a value of 1 in those cases. Thus, the general matching formula becomes min{1, (L-1) / (ML-1)} where ML stands for the level at which the maximum contribution starts. In the case of DMOZ, a reasonable choice for the cut-off point might be level 8 based on the structure in Figure 7.

A disadvantage of the measure just described is the inherent simplifications of using a cut-off point after which all matches are equally 'perfect'. The smaller the cut-off value, the more inaccurate result will become. However, if set too large then some branches of the tree might be too shallow to ever achieve a perfect match. It is furthermore questionable whether a linear measure is appropriate. Hence, we propose the following alternative. The matching scores monotonically increasing from level 1 to 16 (in DMOZ's case) but get close to 1 relatively quickly, say at level 8, and then approaches 1 asymptotically as shown in the figure 11.

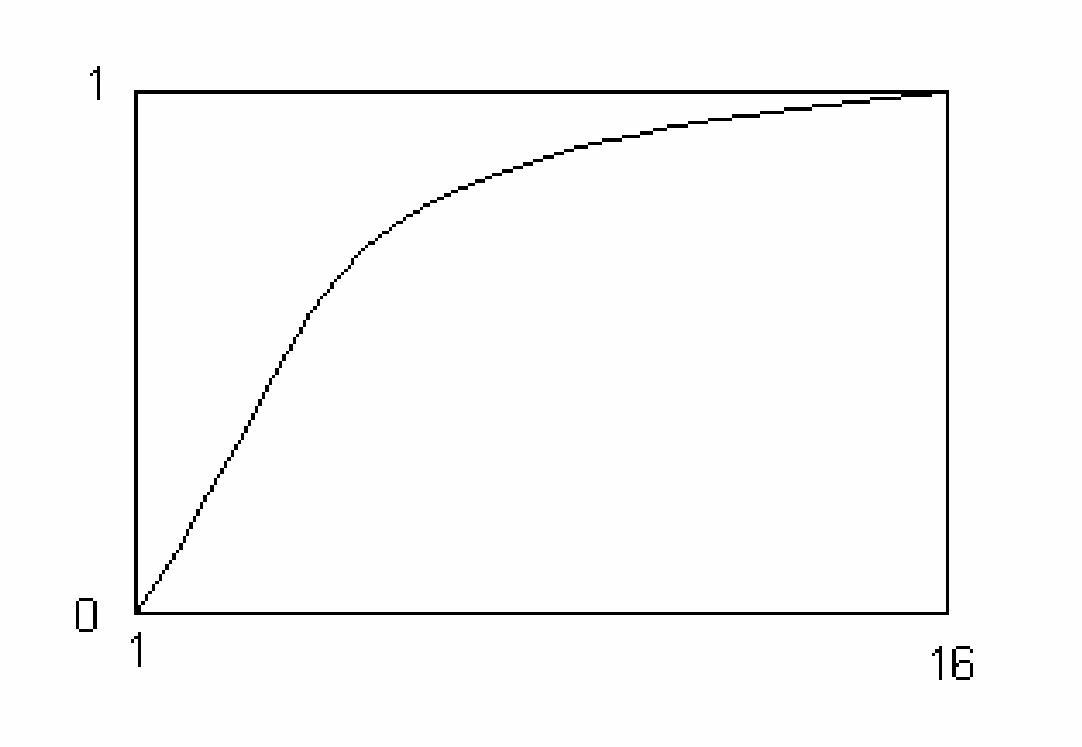

Figure 11: Shape of proposed matching function.

The following equation describes such a function. Let

webprofile1 contain $ca_i$ (i = 1...n ) category addresses

webprofile2 contain $ca_j$ (j = 1…m) category addresses

$ed_{i,j}$ be the edge distance from $ca_i$ to $ca_j$

$l_{i,j}$ be the matching level for $ca_i$ and $ca_j$

Proposed matching function: $-\frac{l_{i,j}^2 - 33l_{i,j} + 32}{240}$

This measure still agrees with the principle that matches at lower levels should score higher but does not unduly penalise the branches that do not go down to the full 16 levels. Assuming we sum the contributions of all the potential matches the total would have to be divided by the total number of matches to transform the metric to a 0 - 1 scale. Hence, the similarity measure $s$ becomes:

$$s = \frac{\sum_{i=1}^{n}\sum_{j=1}^{m}\left(\frac{1}{ed_{i,j}} \times -\left(\frac{l_{i,j}^2 - 33l_{i,j} + 32}{240}\right)\right)}{n \times m}$$

One further factor should be considered when calculating the match between two web profiles. It is the validity of the match if the web profiles have very different numbers of URLs within them (which we will call the disparity correction factor).

If one web profile has only 10 items whilst the other has 100, then a match from these two people would seem to be less valid than one based on web profiles containing 50 and 60 items. This is because in the first case the 10 entries from the first profile have been used proportionately more in calculating the match. Assuming that web profile 1 ($n$ entries) is smaller than web profile 2 ($m$ entries) then finding the fraction $n / m$ would give a higher result to those pairs of profiles which have similar numbers of entries (see column 3 in Figure 12).

However, it would also give a perfect score to two profiles with a very small number of URLs, say 2 URLs each. Clearly, the measure should 'reward' web profiles that have a larger number of entries. One way to do this would be to include the sum of the number of entries. However, some profiles contain a very large number of entries. Analysis of the data shows that users with more than 100 bookmarks are likely to be outliers. Hence, in order to produce a measure in a range from 0 to 1, profiles with more than 100 entries are counted as though they have 100 entries. Column 4 in Figure 12 shows the calculation of such a measure under the assumptions above.

The fifth column in Figure 12 contains the proposed disparity factor. However, if the raw values in column 5 were used the correction effect would probably be stronger than the original matching score. Therefore a scaling parameter $a$ is introduced to reduce the range of the disparity factor. This parameter determines the lowest value in the range ($a$, 1) which the disparity factor can take.

| n | m | n/m | (n+m)/200 | n/m*(n+m)/200 | a+(1-a)*n/m*(n+m)/200 |
|---|---|---|---|---|---|
| 100 | 100 | 1.00 | 1.00 | 1.00 | 1.00 |
| 80 | 100 | 0.80 | 0.90 | 0.72 | 0.89 |
| 60 | 100 | 0.60 | 0.80 | 0.48 | 0.79 |
| 40 | 100 | 0.40 | 0.70 | 0.28 | 0.71 |
| 20 | 100 | 0.20 | 0.60 | 0.12 | 0.65 |
| 80 | 80 | 1.00 | 0.80 | 0.80 | 0.92 |
| 60 | 80 | 0.75 | 0.70 | 0.53 | 0.81 |
| 40 | 80 | 0.50 | 0.60 | 0.30 | 0.72 |
| 20 | 80 | 0.25 | 0.50 | 0.13 | 0.65 |
| 60 | 60 | 1.00 | 0.60 | 0.60 | 0.84 |
| 40 | 60 | 0.67 | 0.50 | 0.33 | 0.73 |
| 20 | 60 | 0.33 | 0.40 | 0.13 | 0.65 |
| 40 | 40 | 1.00 | 0.40 | 0.40 | 0.76 |
| 20 | 40 | 0.50 | 0.30 | 0.15 | 0.66 |
| 20 | 20 | 1.00 | 0.20 | 0.20 | 0.68 |
| 10 | 20 | 0.50 | 0.15 | 0.08 | 0.63 |
| 10 | 10 | 1.00 | 0.10 | 0.10 | 0.64 |
| 5 | 100 | 0.05 | 0.53 | 0.03 | 0.61 |
| 1 | 100 | 0.01 | 0.51 | 0.01 | 0.60 |

Figure 12: Disparity correction using a disparity scaling factor of $a = 0.6$.

Using the same notation as before, with $a$ being the scaling parameter for the disparity correction factor the final similarity measure becomes:

$$s = \frac{\sum_{i=1}^{n}\sum_{j=1}^{m}\left(\frac{1}{ed_{i,j}} \times -\left(\frac{l_{i,j}^2 - 33l_{i,j} + 32}{240}\right) \times (\text{vote } i + \text{vote } j)\right)}{\left(\sum_{i=1}^{n} \text{vote } i + \sum_{j=1}^{m} \text{vote } j\right)} \times \left(a + (1 - a)\frac{n(n+m)}{200m}\right)$$

# 7 CONCLUSIONS

There are a number of steps in the process of preparing the database for use in the Artificial Immune System. These may have an effect on the performance of the system. It will not be possible to tell how critical these issues are until the project is near completion. Having constructed the web profile database the choice of encoding must be made. Again, this could have a critical effect on the success of the Artificial Immune System. It is clear that the construction of a similarity measure that will allow the use of the tree structure is not a trivial task. It may be that this is not necessary and exploration of the potential of the first encoding will be undertaken first since there is already a successful precedent in this case. However, the sparseness of the data set may prevent this, and the creation of a tree comparison similarity measure is an interesting challenge.

To conclude, we believe that with the correct matching metric an idiotypic network based Artificial Immune System should be well suited to supplying interesting yet surprising URLs based on a user's bookmarks. Preliminary results show that with the aid of DMOZ we can map between 60% and 80% of users' bookmarks to votes for suitable categories. We feel confident that this gives us a strong basis for an Artificial Immune System recommender and subsequent result will be published in due course.

### Acknowledgements

The authors would like to thank the many volunteers donating their bookmarks and David Banks for his help with the DMOZ system.